\documentclass{article}

\newtheorem{lemma}{Lemma}
\newcommand{\proof}{\par\penalty-1000\vskip .5 pt\noindent{\bf Proof\/: }}
\newcommand{\blackslug}{\mbox{\hskip 1pt \vrule width 4pt height 8pt 
depth 1.5pt \hskip 1pt}}
\newcommand{\QED}{\quad\blackslug\lower 8.5pt\null\par\noindent}

\begin{document}
\bibliographystyle{plain}
\setlength{\parindent}{0cm}
\setlength{\parskip}{0.5ex}

\title{A note on Darwiche and Pearl}
\author{Daniel Lehmann and students in the Advanced Seminar in AI}
\date{April, 22nd 1998}
\maketitle
\begin{abstract}
It is shown that Darwiche and Pearl's postulates imply 
an interesting property, not noticed by the authors.
\end{abstract}
\section{A short remark}
In~\cite{DarwPearl:AIJ}, Darwiche and Pearl propose postulates for iterated
revisions, noted (R*1) to (R*6) and (C1) to (C4).
In particular, the postulate (C3) reads:
\[
(C3) \ \ \ {\rm If \ } \Psi \circ \alpha \models \mu , {\rm \ then \ }
(\Psi \circ \mu ) \circ \alpha \models \mu.
\]
It will be shown that, in the presence (R*1) to (R*6), (C1) and (C3)
imply:
\[
(**) \ \ \ \ {\rm If \ } \Psi \circ \alpha \models \mu , {\rm \ then \ }
(\Psi \circ \mu ) \circ \alpha \equiv \Psi \circ \alpha.
\]

First, a lemma.
\begin{lemma}
\label{le:ezer}
Assuming (R*1) to (R*6), if \mbox{$\Psi \circ \mu \models \varphi$},
then \mbox{$\Psi \circ \mu \equiv \Psi \circ ( \mu \wedge \varphi )$}.
\end{lemma}
\proof
Since \mbox{$\Psi \circ \mu \models \varphi$},
\mbox{$\Psi \circ \mu \models (\Psi \circ \mu) \wedge \varphi$}.
By (R*4), 
\mbox{$(\Psi \circ \mu) \wedge \varphi \models \Psi \circ (\mu \wedge \varphi)$}.
Therefore \mbox{$\Psi \circ \mu \models \Psi \circ (\mu \wedge \varphi)$}.

If \mbox{$\Psi \circ \mu$} is satisfiable, then,
since \mbox{$\Psi \circ \mu \models \varphi$}, 
\mbox{$(\Psi \circ \mu) \wedge \varphi$} is satisfiable and,
by (R*5), 
\mbox{$\Psi \circ (\mu \wedge \varphi) 
\models (\Psi \circ \mu) \wedge \varphi$}
and therefore \mbox{$\Psi \circ (\mu \wedge \varphi) \models \Psi \circ \mu$}.

If \mbox{$\Psi \circ \mu$} is not satisfiable, then, by (R*3),
$\mu$ is not satisfiable, and \mbox{$ \mu \wedge \varphi$} is not satisfiable.
By (R*1), then, 
\mbox{$\Psi \circ (\mu \wedge \varphi) \models \Psi \circ \mu$}.
\QED
\begin{lemma}
Assuming (R*1) to (R*6), (C1) and (C3), if 
\mbox{$\Psi \circ \alpha \models \mu $}, then
\mbox{$ (\Psi \circ \mu) \circ \alpha \equiv \Psi \circ \alpha$}.
\end{lemma}
\proof
Suppose \mbox{$\Psi \circ \alpha \models \mu$}.
By Lemma~\ref{le:ezer}, 
\mbox{$\Psi \circ \alpha \equiv \Psi \circ (\alpha \wedge \mu)$}.
By (C1), 
\mbox{$\Psi \circ (\alpha \wedge \mu) \equiv 
(\Psi \circ \mu) \circ (\alpha \wedge \mu)$}.
But, by (C3), \mbox{$\Psi \circ \mu \circ \alpha \models \mu$}
and, by Lemma~\ref{le:ezer}, 
\mbox{$\Psi \circ \mu \circ \alpha 
\equiv \Psi \circ \mu \circ (\alpha \wedge \mu)$}.

We conclude that
\mbox{$\Psi \circ \alpha \equiv \Psi \circ \mu \circ \alpha$}.
\QED
\bibliography{/CS/staff/teach/lehmann/ai/papers/my}

\end{document}